\documentclass[runningheads]{llncs}
\usepackage{graphicx}
\usepackage{epstopdf}
\usepackage[misc]{ifsym}
\usepackage{amsfonts}
\usepackage{amsmath}
\usepackage{multirow}
\usepackage{hyperref}
\usepackage{pdfpages}
\hypersetup{hypertex=true,colorlinks=true,linkcolor=blue,anchorcolor=blue,citecolor=blue}

\begin{document}
\title{Multimodal Representations Learning and Adversarial Hypergraph Fusion for Early Alzheimer's Disease Prediction}
%
%

\author{Qiankun Zuo\inst{1} \and
Baiying Lei\inst{2} \and
Yanyan Shen\inst{1} \and
Yong Liu\inst{3} \and
Zhiguang Feng\inst{4} \and
Shuqiang Wang\inst{1}(\textrm{\Letter})}
\institute{Shenzhen Institutes of Advanced Technology, Chinese Academy of Sciences, Shenzhen
	518000, China\\
	\email{\{qk.zuo,yy.shen,sq.wang\}@siat.ac.cn}  \\
	\and Shenzhen University, Shenzhen, 518000, China\\
	\email{leiby@szu.edu.cn}   \\
	\and Renmin University of China, Beijing, 100000, China  \\
	\email{liuyonggsai@ruc.edu.cn}
	\and Harbin Engineering University, Haerbin, 150000 , China\\
	\email{fengzhiguang@hrbeu.edu.cn}
}
%

%
%
%
\maketitle              
\begin{abstract}
Multimodal neuroimage can provide complementary information about the dementia, but small size of complete multimodal data limits the ability in representation learning. Moreover, the data distribution inconsistency from different modalities may lead to ineffective fusion, which fails to sufficiently explore the intra-modal and inter-modal interactions and compromises the disease diagnosis performance. To solve these problems, we proposed a novel multimodal representation learning and adversarial hypergraph fusion (MRL-AHF) framework for Alzheimer's disease diagnosis using complete trimodal images. First, adversarial strategy and pre-trained model are incorporated into the MRL to extract latent representations from multimodal data. Then two hypergraphs are constructed from the latent representations and the adversarial network based on graph convolution is employed to narrow the distribution difference of hyperedge features. Finally, the hyperedge-invariant features are fused for disease prediction by hyperedge convolution. Experiments on the public Alzheimer's Disease Neuroimaging Initiative(ADNI) database demonstrate that our model achieves superior performance on Alzheimer's disease detection compared with other related models and provides a possible way to understand the underlying mechanisms of disorder's progression by analyzing the abnormal brain connections.

\end{abstract}
\section{Introduction}
Alzheimer's disease(AD) is a severe neurodegenerative diseases among the old people and the pathological changes are reflected on the symptoms including memory decline, aphasia and other decreased brain functions~\cite{ref_article1}. Since there is no effective medicine for AD, much attention has been attracted on its prodromal stage, that is, mild cognition impairment(MCI)~\cite{ref_article2}, so that intervention can be implemented to slow down or stop the progression of the disease. With the success of deep learning on medical images analysis~\cite{ref_article2_1,ref_article2_2,ref_article2_3,ref_article2_4,ref_article2_4_1} and other fields~\cite{ref_article2_5,ref_article2_6}, non-invasive magnetic imaging technology becomes an effective tool for detecting dementia at early disease stages, and different modalities carry complementary disease-related information. For example, abnormal functional and structural connectivity between brain regions has been discovered in the resting-state functional magnetic resonance imaging(fMRI)~\cite{ref_article3} and diffusion tensor imaging(DTI)~\cite{ref_article4} modality, respectively; and the T1-weighted magnetic resonance imaging(MRI)~\cite{ref_article5} data contains the information of volume changes in different brain regions. Many researchers~\cite{ref_article6,ref_article7,ref_article8,ref_article9,ref_article9_1} have achieved good performance in brain disease prediction by fusing either two of the above modalities. Therefore, we take all the three modalities as the input of our model to conduct representations learning and fusion for disease diagnosis.

Considering the number of subjects with complete three modalities is limit, it is necessary to make full use of the input data for learning latent representations. The input data can be used to estimate additional distribution, which is prior information for training a much more discriminative and robust model. To make use of additional distribution, the Generative Adversarial Networks(GAN)~\cite{ref_article10} provides an appropriate way for representation learning of the graph data by matching the distribution consistency in representation space. The basic principle is variational inference~\cite{ref_article10a1,ref_article10a2,ref_article10a3} which maximizes the entropy of the probability distribution. It has been applied successfully in medical image analysis~\cite{ref_article10_1,ref_article10_2,ref_article10_3,ref_article10_4} and citation network~\cite{ref_article11,ref_article12}. Besides, Convolution Neural Network(CNN) has great power in recognizing disease-related images~\cite{ref_article13,ref_article13_a1,ref_article13_a2,ref_article13_a4,ref_article13_a5}, which can be utilized to extract features of MRI in data space by a model pre-trained from a great many of unimodal images~\cite{ref_article13_a3,ref_article13_a6}. Therefore, we designed a Distribution-based Graph GAN (D-GraphGAN) and a CNN-based Graph AutoEncoder(CNN-GraphAE) to extract latent representations from fMRI\&DTI and MRI\&DTI, respectively.

After the representations extraction, direct fusion of representations concatenation may lead to bad performance in exploring cross-modal interactions, since the data distributions in representation space may be heterogeneous~\cite{ref_article14}. Adversarial strategy is suitable for translating modality distribution~\cite{ref_article15}. As traditional graph with pairwise regions interaction is not sufficient to characterize the brain network connectivity and fail to encode high-order intra-modal correlations, a hypergraph~\cite{ref_article16} beyond pairwise connections is more suitable to describe the complex brain activities behind dementia. It is found that Hypergraph Neural Networks(HGNN) achieve better performance than Graph Convolutional Networks(GCN) in citation networks~\cite{ref_article17}. Motivated by this, we develop an adversarial hyperedge network to boost multimodal representation fusion performance for AD diagnosis.

In this paper, we propose a Multimodal Representation Learning and Adversarial Hypergraph Fusion(MRL-AHF) to make use of inter-modal complementary and intra-modal correlation information to improve the performance of Alzheimer's disease detection. The estimated additional distribution and pre-trained model are incorporated to improve the ability of representation learning. A hypergraph fusion strategy is adopted to narrow distribution difference in hyperedge space for efficiently fusion by adversarial training. Our MRL-AHN approach is able to enhance the ability of representation learning and boost the multimodal fusing performance. Experiments on the Alzheimer's Disease Neuroimging Initiative(ADNI) database show that our approach achieves superior performance on MCI detection compared with other related works.
\section{Method}
\begin{figure}[h!]
	\includegraphics[width=\textwidth]{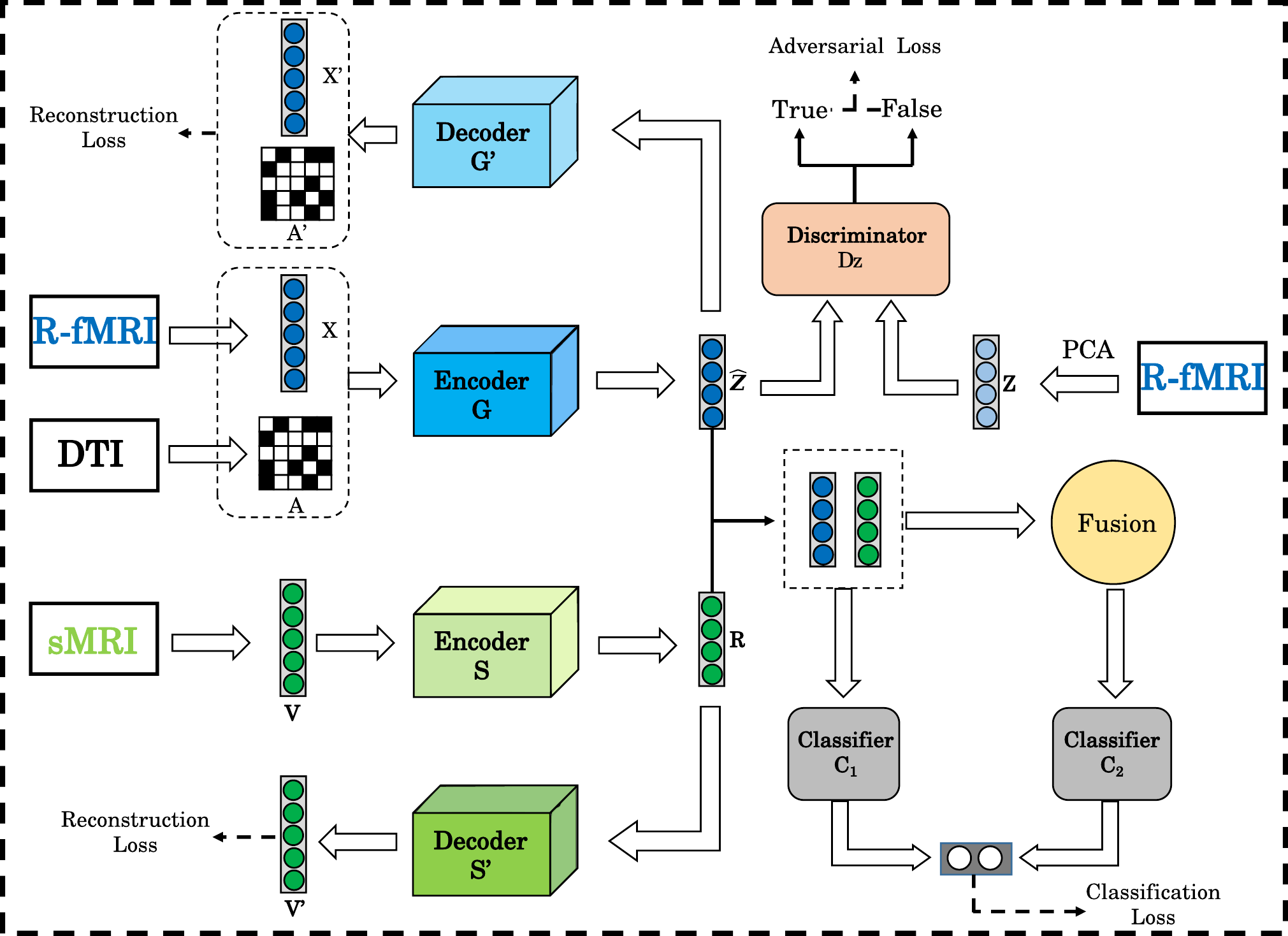}
	\caption{Overall framework of the proposed MRL-AHF for AD diagnosis using fMRI, DTI and MRI data.} \label{fig1}
\end{figure}

An overview of MRL-AHF is given in Fig.~\ref{fig1}. Our framework is comprised of two stages: a representation space learning stage and a adversarial hypergraph fusion stage. The first stage learns the latent representations from fMRI\&DTI and MRI\&DTI by distribution-based GraphGAN and CNN-based GraphAE, respectively. The second stage utilizes the representations output by encoders $G$ and $S$ to conduct hypergraph fusion via adversarial training. The symbol meanings are given bellow: $A$ and $A'$ represent the structural connection(SC) and reconstructed SC matrix, respectively. $X$ and $X'$ denote the functional timeseries(FT) at each brain Region-of-Interest(ROI), the reconstructed FT feature, respectively. $V$ and $V'$ are the feature vector(FV) and reconstructed FV. $\hat{Z}$ and $R$ are features in representation space.

\subsection{Distribution-based GraphGAN}
\subsubsection{Graph construction.} Suppose an indirect graph $\mathcal{G}(\mathcal{V},\mathcal{E})$ is formed with N brain Regions of Interest(ROIs) based on anatomical atlas, $\mathcal{V}=\{\nu_1,\nu_2,...,\nu_N\}$ and $\mathcal{E}=\{\varepsilon_1,\varepsilon_2,...,\varepsilon_N\}$ are a set of nodes and edges, respectively. Specifically, $X=\{x_1,x_2,...,x_N\}\in\mathbb{R}^{N \times d}$ denotes the node feature matrix of brain functional activities derived from fMRI time series, and $A\in\mathbb{R}^{N \times N}$ represents the physical connections matrix reflecting the brain structural information. The element in adjacent matrix $A$ is represented with $A_{ij}=1$ if there exists connection between $i$th and $j$th region, otherwise $A_{ij}=0$.

\subsubsection{Additional distribution estimation.} Normal distribution $\mathbb{N}(0,1)$ cannot represent the graph properly, and an appropriate $Pz$ can boost the ability in learning discriminative representations in adversarial network. In terms of no other known information except for the give graph data $A$ and $X$, we introduce a non-parametric estimation method, Kernel Density Estimation(KDE), to exstimate $Pz(Z|X)$ that approximates to $Pz(Z|X,A)$ by combining both anatomical and nueroimaging information. Specifically, based on some certain disease-related ROIs, we can obtain a set of nodes $U\subseteq\mathcal{V}$ by applying Determinant Point Process(DPP)~\cite{ref_article18} method on matrix $A$, and the corresponding node features are selected to form features matrix $X_U\in\mathbb{R}^{m \times d}$ with $m=|U|$ nodes, followed with dimension reduction by Principal Component Analysis(PCA) to get $Z_U\in\mathbb{R}^{m \times q}$. $q$ is the dimension in latent representation space. Assuming $Z_i$ is a latent representation of each node, $Pz(Z)$ can be defined by
\begin{equation}
Pz(Z) \approx \frac{1}{mb} \sum_{i=1}^{m} K(\frac{Z-Z_i}{b})
\end{equation}
Where $K(\cdot)$ is a multi-dimensional Gaussian Kernel function, $b$ denotes the bandwidth that determines the smoothness of the distribution

\subsubsection{GraphGAN.}The encoder $G$ encodes $A$ and $X$ as latent representations $\widehat{Z}$, which are sent to the discriminator $D_{Z}$ as negative samples. The positive samples $Z$ are estimated from the additional distribution $Pz(Z|X,A)$. The adversarial loss function is defined as follows
\begin{equation}
\pounds_{D_{Z}} = -\mathbb{E}_{A \sim P_A, X \sim P_X}[D_{Z}(G(A,X))] + \mathbb{E}_{{Z} \sim P_Z}[D_{Z}({Z})]
\end{equation}
\begin{equation}
\pounds_{G} = \mathbb{E}_{A \sim P_A, X \sim P_X}[D_{Z}(G(A,X))]
\end{equation}

Besides, the reconstruction loss and the classification loss are given bellow:
\begin{equation}
\pounds_{Rec1} = \mathbb{E}_{A \sim P_A, X \sim P_X}[f(X,X')] + \mathbb{E}_{A \sim P_A, X \sim P_X}[f(A,A')]
\end{equation}
\begin{equation}
\pounds_{Cls1} = \mathbb{E}_{A \sim P_A, X \sim P_X}[y \cdot logy']
\end{equation}
Where, $X'=G'(A,\hat{Z})$, $\hat{Z}=G(A,X)$ and $ A' = \sigma(\hat{Z} \hat{Z}^T)$ are the reconstructed graph data, $f(a,b)=a \cdot logb + (1-a) \cdot log(1-b)$ is binary cross entropy function, $y'=C_1(\hat{Z})$ is the predicted labels. $C_1$ is a two-layer perception. $G$ and $G'$ are two-layer GCN, specifically.

\subsection{CNN-based GraphAE}
A dense convolutional network with 4 blocks is trained on large number of labeled images and then used to extract a feature vector $V$ for each MRI using the last fully connected layer.  In order to deploy CNN feature on the ROIs, we equally distribute the feature $V\in\mathbb{R}^{1 \times 128}$ on the ROIs, and the SC is used to guide the feature to flow between two connected nodes. The latent representations $R$ is obtained by a two-layer GCN Encoder $S$, followed with a decoder $S'$ to reconstruct features. The reconstruct loss and classification loss are defined as
\begin{equation}
\pounds_{Rec2} = \mathbb{E}_{V \sim P_V}[f(V,V')]
\end{equation}
\begin{equation}
\pounds_{Cls2} = \mathbb{E}_{V \sim P_V}[y \cdot logy'']
\end{equation}
Where, $V'=S'(R)=S'(S(V))$, $y$ is the truth one-hot label, $y''=C_1(R)=C_1(S(V))$  is the predicted label.

\subsection{Adversarial Hypergraph Fusion}
\subsubsection{Hypergraph construction.} By denoting a hyperedge $E$ connecting multiple nodes, we can construct a hyperedge for each node centered. Specifically, we use K-NearestNeighbor(KNN) method to select the nodes for each hyperedge based on the Euclidean distance. At last, we can get an incident matrix $H_1$ and $H_2$ from the learned representations $\hat{Z}$ and $R$, respectively. The formula is given as follows:
\begin{equation}
H(\mathcal{V},E)=\{_{1,if\   \nu \notin E}^{0, if\   \nu \in E}
\end{equation}

\subsubsection{Adversarial Hypergraph Learning.} In order to narrow the gap between the representations, we adopt the adversarial training strategy to make the distribution from different modalities the same. The hypergraph convolution is spitted into convex convolution and hyperedge convolution.  The hyperedge feature of $\hat{Z}_H$ is computed by Vertex aggregation is sent to the discriminator $D_H$ as a positive sample, the negative sample $R_H$ is obtained by Vertex convolution of hypergraph $R$, the formula is illustrated as follows:
\begin{equation}
\hat{Z}_H=D_{1e}^{-1/2} H_1^T D_{1e}^{-1/2} \hat{Z}
\end{equation}
\begin{equation}
R_H=D_{2e}^{-1/2} H_2^T D_{2e}^{-1/2} R \Theta
\end{equation}
Where, $D_{1e}$ and $D_{2e}$ are the edge degree of $H_1$ and $H_2$, respectively; $\Theta$ is the weighting parameters. Both $\hat{Z}_H$ and $R_H$ are sent to the discrminatr $D_H$ for adversarial training.Then, we fuse the hyperedge features by using edge aggregation to get vertex feature matrix $F$ as follows
\begin{equation}
F=D_{1v}^{-1/2} H_1 D_{1v}^{-1/2} \hat{Z}_H + D_{2v}^{-1/2} H_2 D_{2v}^{-1/2} R_H
\end{equation}

Finally, the fused features is used to construct connectivity matrix by bilinear pooling and  then sent to the classifier for task learning. The adversarial and classification loss are given below
\begin{equation}
\begin{split}
\pounds_{AHF} &= \pounds_{D_H} + 0.1\pounds_{Ver} + \pounds_{Cls3} \\
&= \mathbb{E}_{A \sim P_A, Z \sim P_Z}[D_H(\hat{Z}_H)] -0.1\cdot\mathbb{E}_{V \sim P_V}[D_H(R_H)] \\
& \ \ \  + \mathbb{E}_{A \sim P_A, Z \sim P_Z, V \sim P_V}[y \cdot logy''']
\end{split}
\end{equation}
Here, $D_{1v}$ and $D_{2v}$ are the node degree of $H_1$ and $H_2$, respectively; and $y'''=C_2(\sigma(FF^T))$ is the predicted label.

\subsection{Training strategy}
In Conclusion, the total loss of the proposed frame is:
\begin{equation}
\begin{split}
\pounds_{MRLAHF}= \pounds_{G} + 0.1 \pounds_{D_{Z}} + \pounds_{Rec1} + \pounds_{Rec2} + \pounds_{Cls1}+\pounds_{Cls2}  + \gamma \pounds_{AHF}
\end{split}
\end{equation}
Where $\gamma$ is a hyper-parameter that determines the relative importance of feature fusion loss items.

During the training process, firstly, we update the generators, encoders and decoders with the loss backpropogation of $\pounds_{Rec1}$, $\pounds_{Rec2}$ and $\pounds_{G}$; next, we use the $\pounds_{D_{Z}}$ to update the discriminator to improve the discriminator ability of joint and marginal distribution; then, $\pounds_{Cls1}$ and $\pounds_{Cls2}$ are utilized to update encoders and classifier to boost the performance of task learning. After the discriminative representations have been extracted, $\pounds_{D_H}$ and $\pounds_{Ver}$ are performed to update the parameters in vertex convolution and discriminator $D_H$ alternatively; finally, $\pounds_{Cls3}$ updates the classifier $C_2$ to get a discrminative decision on the fused features.

\section{Experiments}
\subsection{Data}
A total of 300 subjects from ADNI database are used for this study with complete three modalities: fMRI, DTI and T1-weighted MRI, including 64 AD patients (39 male and 25 female, mean age 74.7, standard deviation 7.6), 76 late MCI patients (43 male and 33 female, mean age 75.8, standard deviation 6.4), 82 early MCI patients (40 male and 42 female, mean age 75.9, standard deviation 7.5), and 78 normal controls (39 male and 39 female, mean age 76.0, standard deviation 8.0).

For T1-weighted MRI data, we follow the standard proprocessing steps, including  strip non-brain tissue of the whole head, image reorientation, resampling into a voxel size of 91x109x91 in Neuroimaging Informatics Technology Initiative(NIFTI) file format and extracting a 128-dimensional feature vector FV by a pre-trained densnet model. The fMRI data is preprocessed using GRETNA toolbox to obtain node features FT with a size of 90x187, the main steps include magnetization equilibrium, head-motion artifacts, spatial normalization, spatial filter with 0.01-0.08Hz, regression of local time-series, warping automated anatomical labeling(AAL90) atlas and removing the first 10 timepoints. The DTI data preprocessing operation is performed using PANDA toolbox to get 90x90 matrix SC. The detailed procedures are skull stripping, resolution resampling, eddy currents correction, fiber tracking. The generated structural connectivity is input to our model as graph structure.

\subsection{Experimental settings}
In this study, we use three kinds of binary classification task, i.e., (1)EMCI vs. NC; (2)LMCI vs. NC; (3)AD vs. NC. 10-fold cross validation is selected for task learning. In order to demonstrate the superiority of our proposed model compared with other models, we introduce previous methods for comparison. (1) Support Vector Machine(SVM)~\cite{ref_article18_00}; (2) two layers of the diffusion convolutional neural networks(DCNN)~\cite{ref_article18_01}; (3) our method with only fMRI and DTI; (4) our method with complete three modalities. For convenient viewing, the above methods using fMRI and DTI are denoted SVM(F-D), DCNN(F-D) and Ours(F-D).

In the experiments, we set the model parameters as follows:$N=90$, $m=10$, $q=32$, $\gamma=0.5$, tanh and sigmoid activation function for generators and decoders, respectively. The disease-related ROIs are selected according to previous studies~\cite{ref_article2,ref_article8,ref_article9}. $C_1$ is a two-layer perception with 16-neuron and 2-neuron in the hidden and output layers. $C_1$ is a two-layer perception with 90-neuron and 2-neuron in the hidden and output layers. $G$ and $G'$ are two-layer GCN, specifically, the hidden and output layers of $G$ is 64-neuron and 32-neuron, while the hidden and output layers of $G'$ is 64-neuron and 187-neuron. The hidden and output layers of $S$ is 64-neuron and 32-neuron, the hidden and output layers of $S'$ is 64-neuron and 128-neuron. For the discriminator $D_{Z}$, the hidden layer contains 1 filter with the size 32x1, the output layer contains 1 filters with the size 90x1. For the discriminator $D_H$, the filter size of hidden layer is 1x90, the filter size of output layer is 90x1. To balance the adversarial training, we choose 0.001 learning rate for the generators, encoders, decoders, and classifiers, 0.0001 learning rate for the discriminators. In the training process, 100 epochs are employed on representation learning, followed with 200 epochs for adversarial hypergraph fusion.

\subsection{Results}
Table~\ref{tab1} summarizes the results of different methods in three binary classification tasks using 10-fold cross validation. As can be seen that our proposed model has the best accuracy of 95.07\%, 91.56\%,  and 87.50\% in the tasks of AD vs. NC, LMCI vs. NC, EMCI vs. NC, respectively. Our method behaves better than other methods. It is found that introducing more modal images is beneficial to improve model detection performance. In addition, comparisons between ours(F-D) and DCNN(F-D) indicate that adding distribution-guided GraphGAN can improve detection accuracy. What's more, when compared with other related algorithms as illustrated in table~\ref{tab2}, the proposed method has achieved superior performance for MCI detection. It ourperforms the literature by 2.08\%. Note that, methods using fMRI and DTI are denoted  F-D, and Ours mean the proposed model using three complete modalities.

\begin{table}
	\caption{Mean detection performance of the proposed and related methods.(\%)}\label{tab1}
	\begin{center}
		\resizebox{\textwidth}{12mm}{
			\begin{tabular}{c|cccc|cccc|cccc}
				\hline \cline{1-13}
				\multicolumn{1}{c|}{\multirow{2}{*}{Method}} & \multicolumn{4}{|c|}{AD vs. NC} & \multicolumn{4}{|c|}{LMCI vs. NC} & \multicolumn{4}{|c}{EMCI vs. NC}\\ \cline{2-13}
				\multicolumn{1}{c|}{} & Acc	& Sen & Spec & Auc & Acc & Sen & Spec & Auc & Acc & Sen & Spec & Auc\\
				\hline
				SVM(F-D) & 76.05 & 70.31 & 80.76 & 83.16 & 69.48 & 64.47 & 74.35 & 78.23 & 65.62 & 54.87 & 76.92 & 71.46\\
				\hline
				DCNN(F-D) & 84.51 & 87.50 & 82.05 & 89.40 & 79.87 & 77.63 & 82.05 & 84.29 & 76.25 & 76.83 & 75.64 & 82.68\\
				\hline
				Ours(F-D) & 88.73 & 84.37 & 92.31 & 97.48 & 84.42 & 84.21 & 84.62 & 94.01 & 82.50 & 82.93 & 82.05 & 91.65\\
				\hline
				\bfseries Ours & {\bfseries 95.07} & {\bfseries 93.75} & {\bfseries 96.15} & {\bfseries 98.20} & {\bfseries 91.56} & {\bfseries 94.74} & {\bfseries 88.89} & {\bfseries 94.64} & {\bfseries 87.50} & {\bfseries 86.59} & {\bfseries 88.46} & {\bfseries 93.05}\\
				\hline
		\end{tabular}}
	\end{center}
\end{table}

\begin{table}
	\caption{Algorithm comparison with the related works.(\%)}\label{tab2}
	\begin{center}
		\begin{tabular}{c|c|c|cccc}
				\hline \cline{1-7}
				\multicolumn{1}{c|}{\multirow{2}{*}{Method}} & \multicolumn{1}{|c|}{\multirow{2}{*}{Modality}} & \multicolumn{1}{|c|}{\multirow{2}{*}{subject}} & \multicolumn{4}{|c}{MCI vs. NC}\\ \cline{4-7}
				\multicolumn{1}{c|}{} & \multicolumn{1}{|c|}{} & \multicolumn{1}{|c|}{} & Acc	& Sen & Spec & Auc\\
				\hline
				Xing et al.~\cite{ref_article7} & fMRI\&MRI & 368 & 79.73 & 86.49 & 72.97 & -\\
				\hline
				Yu et al.~\cite{ref_article8} & fMRI\&DTI & 184 & 85.42 & 86.57 & 84.42 & 89.98\\
				\hline
				Zhu et al.~\cite{ref_article18_1} & MRI\&PET\&CSF & 152 & 83.54 & 95.00 & 62.86 & 78.15\\
				\hline
				\bfseries Ours & MRI\&fMRI\&DTI & 160 & {\bfseries 87.50} &  86.59 & {\bfseries 88.86} & {\bfseries 93.05}\\
				\hline
		\end{tabular}
	\end{center}
\end{table}

\begin{figure}
	\begin{center}
		\includegraphics[width=\textwidth]{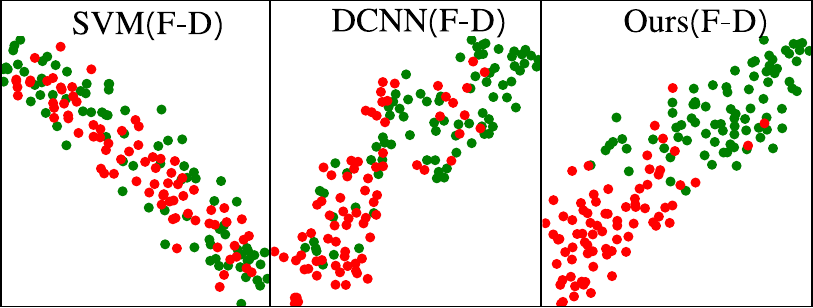}
		\caption{Visualization of features of SVM, DCNN and our method using t-SNE tools for EMCI vs. NC. F-D means the three methods are input with fMRI and DTI.} \label{fig2}
	\end{center}
\end{figure}

We further investigate the classification performance of our model by t-SNE analysis. Fig.~\ref{fig2} shows the projection of features of three methods on the two dimensional plane in different task learning. Our model has slim and easily divisible plane compared with SVM and DCNN, indicating that the feature obtained from our method is more discriminative than that of SVM or DCNN. This investigation explains in detail why our model performs better than others in task learning.

\begin{figure}
	\begin{center}
		\includegraphics[width=\textwidth]{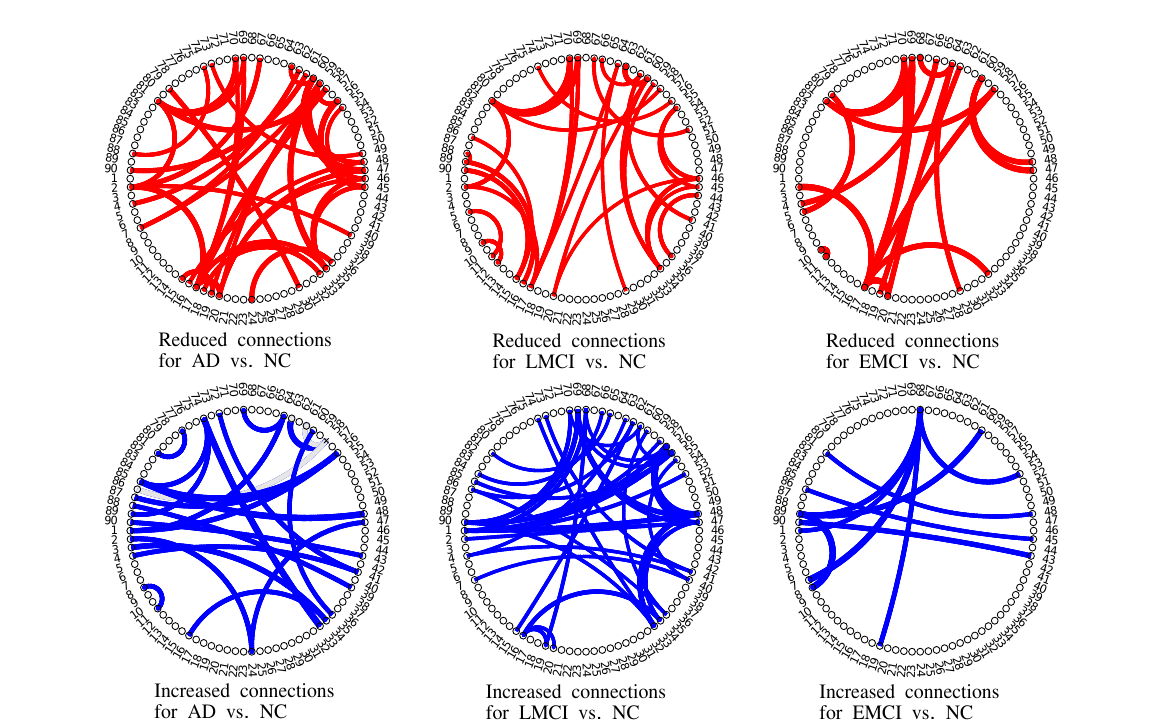}
		\caption{Visualization of brain connection changes at different stages.} \label{fig3}
	\end{center}
\end{figure}

Since the interactions among multiple regions are beneficial for characterizing disease-related brain activities, we construct connectivity matrix using the fused features by bilinear pooling. As is displayed in Fig.~\ref{fig3}, we mean the connectivity matrices of each group for each binary classification task and then subtract patients connectivity matrix from NC connectivity matrix to obtain the change of brain network connections. It gives the following information: the connections gradually reduce as the disease worsens, while the increased connections rise up in early stages and drop to a low level when deteriorated to AD. This phenomenon may be explained by compensatory mechanism generation and weakening in the progression of MCI to AD~\cite{ref_article19,ref_article20,ref_article21}.

\section{Conclusion}
In this paper, we proposed a novel Multimodal-Representaion-Learning and Adversarial-Hypergraph-Fusion frame work for Alzheimer's disease diagnosis. Specifically, features in representations space are extracted by distribution-based GraphGAN and CNN-based GraphAE, respectively. And an adversarial strategy in modal fusion is utilized for AD detection. Results on ADNI dataset demonstrate that prior information can help to enhance discrimination of representation learning and adding more modalities can boost the detection performance. Furthermore, The study on multimodal fusion gives a possible way to understand the disorder's underlying mechanisms by analyzing the abnormal brain connections. In our future work, we will focus the abnormal connections among some certain ROIs and extend this work to multitask classification.

\subsubsection*{Acknowledgment.}
This work was supported by the National Natural Science Foundations of China under Grant 61872351, the International Science and Technology Cooperation Projects of Guangdong under Grant 2019A050510030, the Distinguished Young Scholars Fund of Guangdong  under Grant 2021B1515020019, the Excellent Young Scholars of Shenzhen under Grant RCYX20200714114641211 and  Shenzhen Key Basic Research Project under Grant JCYJ20200109115641762.

%
%
%
%

\end{document}